\documentclass[letterpaper]{article} 
\usepackage[draft]{aaai2026}  
\usepackage{times}  
\usepackage{helvet}  
\usepackage{courier}  
\usepackage[hyphens]{url}  
\usepackage{graphicx} 
\usepackage{natbib}  
\usepackage{caption} 
\DeclareUnicodeCharacter{266B}{\textmusicalnote}
\usepackage{algorithm}

\usepackage{algpseudocode}
\usepackage{graphicx}
\usepackage{textcomp}
\usepackage{xcolor}
\usepackage{booktabs}
\usepackage{multirow}
\usepackage{multicol}
\usepackage{natbib}
\usepackage{tikz}
\usepackage{pifont}
\usepackage[most]{tcolorbox}
\usepackage{xcolor}
\usepackage{enumitem}

\usepackage{todonotes}

\usepackage{newfloat}
\usepackage{listings}
\DeclareCaptionStyle{ruled}{labelfont=normalfont,labelsep=colon,strut=off} 
\floatstyle{ruled}
\newfloat{listing}{tb}{lst}{}
\floatname{listing}{Listing}
\title{You Don't Need Pre-built Graphs for RAG: \\ Retrieval Augmented Generation with Adaptive Reasoning Structures}
\author{
    Shengyuan Chen\thanks{Equal contribution. $^{\dagger}$Corresponding author.},
    Chuang Zhou$^*$, 
    Zheng Yuan,
    Qinggang Zhang$^{\dagger}$,
    Zeyang Cui,
    Hao Chen,
    Yilin Xiao,
    Jiannong Cao,
    Xiao Huang
}
\affiliations{
    The Hong Kong Polytechnic University\\
    \{sheng-yuan.chen, qinggang.zhang, jiannong.cao, xiao.huang\}@polyu.edu.hk, \\ \{chuang-qgzj, yzheng.yuan, ze-yang.cui, yilin.xiao\}@connect.polyu.hk, sundaychenhao@gmail.com
}

\usepackage{bibentry}

\begin{document}

\maketitle

\begin{abstract}
Large language models (LLMs) often suffer from hallucination, generating factually incorrect statements when handling questions beyond their knowledge and perception. Retrieval-augmented generation (RAG) addresses this by retrieving query-relevant contexts from knowledge bases to support LLM reasoning. Recent advances leverage pre-constructed graphs to capture the relational connections among distributed documents, showing remarkable performance in complex tasks. However, existing Graph-based RAG (GraphRAG) methods rely on a costly process to transform the corpus into a graph, introducing overwhelming token cost and update latency. Moreover, real-world queries vary in type and complexity, requiring different logic structures for accurate reasoning. The pre-built graph may not align with these required structures, resulting in ineffective knowledge retrieval.
To this end, we propose a \textbf{\underline{Logic}}-aware \textbf{\underline{R}}etrieval-\textbf{\underline{A}}ugmented \textbf{\underline{G}}eneration framework (\textbf{LogicRAG}) that 
dynamically extracts reasoning structures at inference time to guide adaptive retrieval without any pre-built graph. 
LogicRAG begins by decomposing the input query
into a set of subproblems and constructing a directed acyclic
graph to model the logical dependencies among
them. To support coherent multi-step reasoning, LogicRAG then lin-
earizes the graph using topological sort, so that subproblems
can be addressed in a logically consistent order.  Besides, LogicRAG applies graph pruning to reduce redundant retrieval and uses context pruning to filter irrelevant context, significantly reducing the overall token cost.
Extensive experiments demonstrate that LogicRAG achieves both superior performance and efficiency compared to state-of-the-art baselines.

\end{abstract}

\begin{links}
    \link{Code}{https://github.com/chensyCN/LogicRAG}
    \link{Extended version}{https://arxiv.org/abs/2508.06105}
\end{links}

\section{Introduction}

Large language models (LLMs), like Claude~\cite{anthropic2024claude} and ChatGPT~\citep{openai2023gpt4}, have shown remarkable ability in a wide range of tasks like complex reasoning~\citep{LLM4EA, zhou2025self, zhou2025each, hong2024knowledge, hong2025next}, question answering~\citep{khashabi2020unifiedqa}, and social analysis~\citep{lu2021engage}. However, these foundation models frequently struggle with domain-specific tasks, often producing hallucinations or inaccurate responses when handling knowledge-intensive queries~\citep{huang2023survey,fang2025safemlrm,fang2024alphaedit}. To mitigate these issues, retrieval-augmented generation (RAG)~\citep{gao2023retrieval,lewis2020retrieval,zhang2025erarag} emerges as a promising framework that enhances LLMs by retrieving query-relevant contexts from external knowledge bases.

\begin{figure}
    \centering
    \includegraphics[width=\linewidth]{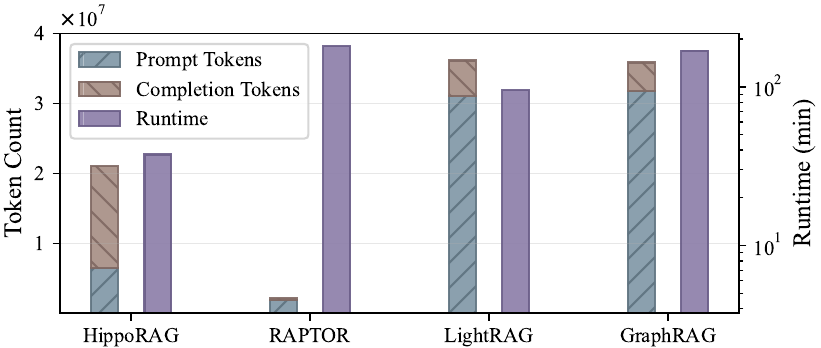}
    \caption{Token and runtime cost of the graph construction process of graph-based RAG methods on 2WikiMQA.}
    \label{fig:graph-construction-cost}
    \vspace{-1em}
\end{figure}

Real-world RAG systems often face significant challenges when handling large-scale, unstructured domain corpora~\citep{peng2024graph,zhang2025faithfulrag}. Documents sourced from research papers, textbooks, or technical reports vary widely in reliability and completeness~\citep{guo2025empowering, zhong2024unrealzoo, wu2025hierarchical}, and the retrieved information is often complex and disorganized, as domain knowledge is typically scattered across multiple sources without clear dependencies~\citep{sun2024thinkongraph,ma2024think-on-graph-2.0}. To manage this complexity, RAG systems commonly segment documents into smaller chunks for indexing~\citep{borgeaud2022improving,izacard2023atlas,jiang2023active}. This approach, however, sacrifices critical contextual information, leading to reduced retrieval accuracy and limited capability for complex reasoning tasks, particularly those requiring multi-hop reasoning across interconnected concepts.

To address this, recent advances~\cite{zhang2024knowgpt,xiang2025use,li2023survey,zhou2025depth} leverage pre-constructed graphs to capture the relational connections among distributed documents, showing remarkable performance in complex tasks. 
The typical graph-based RAG (GraphRAG) approaches like Microsoft GraphRAG~\citep{edge2024local} utilize hierarchical community-based search and combine local and global querying to enhance response quality. Similarly HippoRAG~\citep{HippoRAG} leverages PageRank-inspired algorithms to prioritize highly relevant nodes in memory-augmented retrieval for enhanced contextual coherence, 
RAPTOR~\citep{sarthi2024raptor} uses recursive abstractive processing for hierarchical text representation, and KGP~\citep{wang2024knowledge} constructs domain-specific knowledge graphs with probabilistic reasoning to reduce noise. LinearRAG~\citep{linearrag} explores relation-free graph construction, eliminating LLM token costs and improve robust graphRAG with elegant bipartite graphs.
These GraphRAG methods excel in handling multi-hop queries by providing structured contextual depth.

\begin{figure*}[htbp]
    \centering
\includegraphics[width=\linewidth]{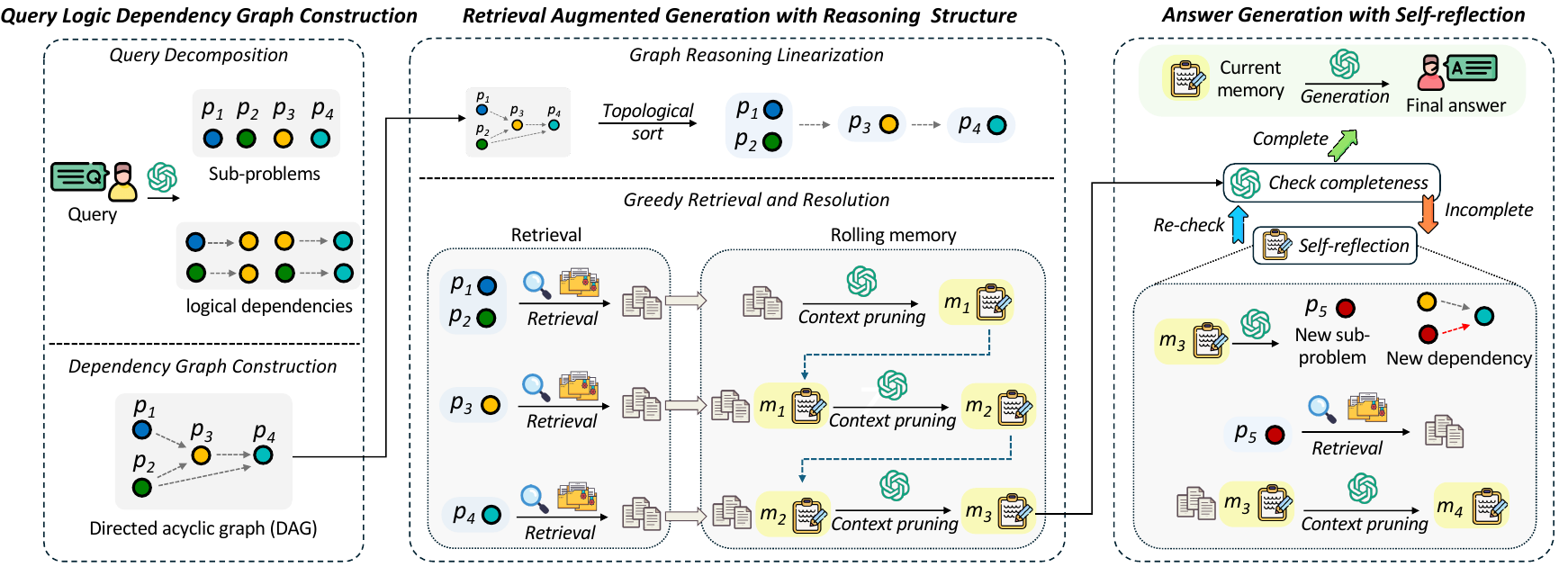}
    \caption{Illustration of the proposed LogicRAG.}
    \label{fig:LogicRAG}
\end{figure*}

Despite recent advances, GraphRAG systems still face critical limitations in real-world scenarios. (i) Efficiency issues. Existing GraphRAG models rely on a costly process to transform the corpus into a graph, introducing overwhelming token cost and update latency as shown in Figure~\ref{fig:graph-construction-cost}. It is hard to generalize to practical scenarios where knowledge bases are large-scale or dynamically evolving ~\citep{edge2024local}.
(ii) Low quality of the pre-built graph. Existing methods leverage LLMs to automatically build the graph without any guidance, which may introduce irrelevant or redundant information, leading to inefficiencies in both retrieval and reasoning~\citep{guo2024lightrag}. 
(iii) Lack of flexibility. Real-world queries vary in type and complexity, requiring different logic structures for accurate reasoning~\citep{peng2024graph}. The pre-built graph may not align with these required structures, resulting in ineffective knowledge retrieval. These challenges highlight the need for a more adaptive and efficient approach. 

To this end, we propose LogicRAG, which dynamically extracts reasoning structures at inference time to guide adaptive retrieval without any pre-constructed graph.
Specifically, LogicRAG begins by decomposing the input query into a set of subproblems and constructing a directed acyclic graph (DAG) to model the logical dependencies among them. This structured representation enables adaptive planning of the retrieval process by identifying which evidence chunks are logically connected to each subproblem. To support coherent multi-step reasoning, LogicRAG then linearizes the graph using topological sort, so that subproblems can be addressed in a logically consistent order. To further improve efficiency without compromising performance, the model applies graph pruning to reduce redundant retrieval and uses context pruning to filter irrelevant context, significantly reducing the overall token cost.

Our contributions are summarized as follows:
\begin{itemize}
    \item We identify the limitations of pre-built graphs used in existing GraphRAG models, and propose LogicRAG, a novel framework that dynamically extracts reasoning structures at inference time to guide adaptive retrieval without any pre-built graph.
    \item We model query logical dependencies with a directed acyclic graph, offering a principled and universal modeling of reasoning structure.
    \item We enable efficient reasoning by \textit{graph reasoning linearization} and \textit{context and graph pruning}.
    \item Extensive experiments on benchmark datasets show that  LogicRAG achieves both superior performance and efficiency compared to state-of-the-art baselines.
\end{itemize}

\section{Problem Statement}

Given an input query \( Q \), retrieval-augmented generation aims to produce an answer \( A \) by retrieving relevant context \( \mathcal{C} \) from an external knowledge corpus \( \mathcal{K} \) and generating a response using a language model. Formally:
\[
A = f_{\mathrm{RAG}}(Q, \mathcal{C}), \quad \mathcal{C} = \mathcal{R}(Q).
\]
The key in a RAG system is the relevance of the retrieved context to the query. While direct semantic matching is often sufficient for simple fact retrieval tasks, it becomes inadequate for complex queries that require assembling multiple pieces of supporting knowledge. These pieces are often logically interconnected and must be retrieved and composed in a way that supports coherent reasoning. As such, the problem is to identify and retrieve a structured set of relevant contexts from \( \mathcal{K} \) that collectively address the query's underlying information needs. This requires reasoning over both the content and the relationships among the retrieved evidence to guide the generation model toward producing accurate, complete, and logically consistent answers.


\section{The Framework of LogicRAG}

\subsection{Overview}

To handle complex queries in RAG, we propose a structured inference framework that decomposes complex queries into interdependent subproblems and resolves them via a logic-guided retrieval and generation process. The core of our method is the construction and utilization of a Query Logic Dependency Graph, a directed acyclic graph (DAG) that models the logical structure underlying the query. Each node in the DAG represents a subproblem, while edges encode the directional dependencies required for reasoning.

The framework operates in three sequential stages. First, the input query is decomposed into subproblems, and a DAG is constructed to capture their logical relationships. This graph is dynamically adapted during inference to reflect evolving retrieval needs. Second, the DAG is topologically sorted to produce a linear execution order that respects the dependencies among subproblems. Each subproblem is then resolved in a greedy, forward-pass manner, wherein retrieval is conditioned on the outputs of previously resolved subproblems. This process ensures context-aware retrieval and avoids recursive dependencies that hinder efficiency. Finally, to enhance scalability, we apply a two-dimensional pruning strategy that reduces context redundancy and merges semantically similar subproblems. Context pruning uses LLM-based summarization to maintain a rolling memory of relevant information, while graph pruning consolidates loosely coupled subproblems for unified resolution.

This logic-aware RAG pipeline transforms the traditionally flat retrieval paradigm into a dependency-sensitive inference mechanism. By aligning retrieval operations with the query’s internal reasoning structure, the framework enables efficient, accurate, and scalable multi-step reasoning over complex information needs.






\subsection{Query Logic Dependency Graph Construction}

To model the interdependencies among sub-problems during inference, we construct a logic dependency graph that adaptively captures the structure of complex queries. This graph is instantiated as a directed acyclic graph (DAG), which guides the retrieval process in a manner aligned with the query's reasoning structure. We discuss the details of the graph modeling and graph construction below:

\textbf{Graph modeling.} Let \( Q \) denote the input query, decomposed into a set of subproblems \( P = \{p_1, p_2, \dots, p_n\} \), where each \( p_i \) corresponds to a node \( v_i \) in a DAG \( G = (V, E) \). The node set \( V = \{v_1, v_2, \dots, v_n\} \) represents the subproblems, and the directed edge set \( E \subseteq V \times V \) encodes logical dependencies among them. The acyclic nature of \( G \) ensures a well-defined ordering for retrieval and reasoning.

\textbf{Graph construction.} 
The DAG construction process is guided by three key considerations:  
\ding{182} \textit{Decomposition Accuracy}: The query \( Q \) is first segmented into subproblems via LLM-based reasoning, using few-shot prompting to ensure precise task separation.  
\ding{183} \textit{Dependency Modeling}: Edges are inferred by the LLM based on logical precedence among subproblems, and the resulting graph is verified via topological sorting to enforce acyclicity.  
\ding{184} \textit{Dynamic Adaptation}: If retrieval for a node yields insufficient context, the LLM dynamically augments the DAG by adding new subproblems and updating dependencies.

This DAG-based representation enables a principled representation of the query structures by explicit modeling of conditional dependencies across subproblems, supporting query decomposition and prioritization of retrieval. By aligning retrieval with the query's internal logic, the model is better equipped for complex reasoning.


\begin{figure*}[t]
    \centering
    \begin{minipage}{0.65\linewidth}
        \centering
        \includegraphics[width=\linewidth]{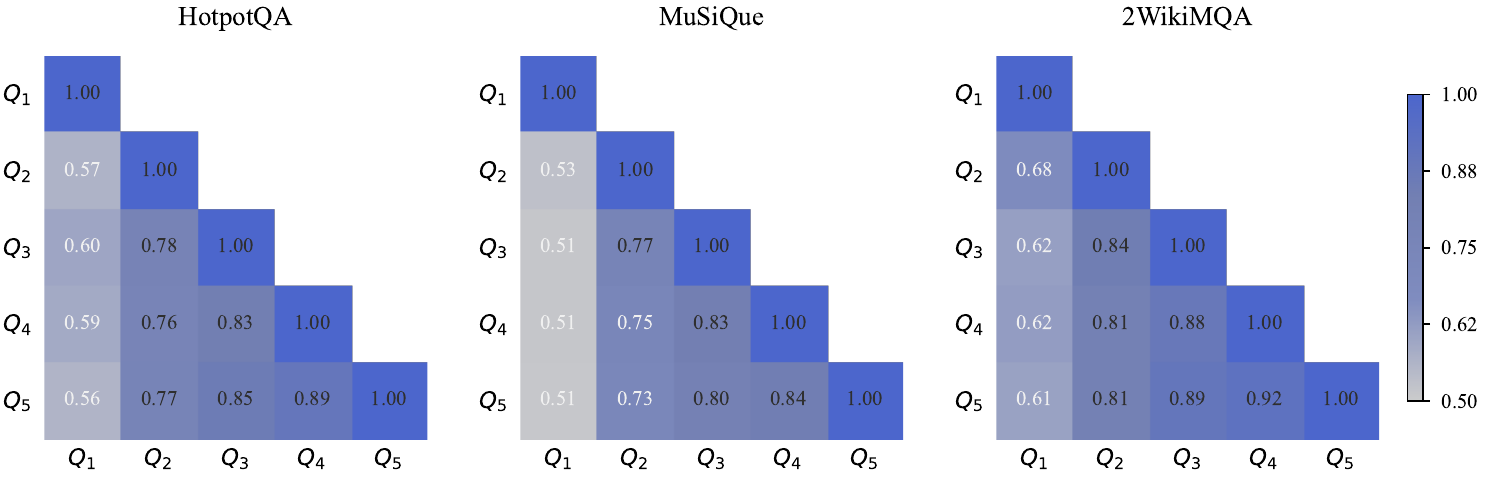}
        \caption{Word-level Jaccard similarity between subqueries across rounds in the agentic RAG process, averaging across the dataset.}
        \label{fig:sq_sim}
    \end{minipage}%
    \hfill
    \begin{minipage}{0.33\linewidth}
        \centering
        \includegraphics[width=\linewidth]{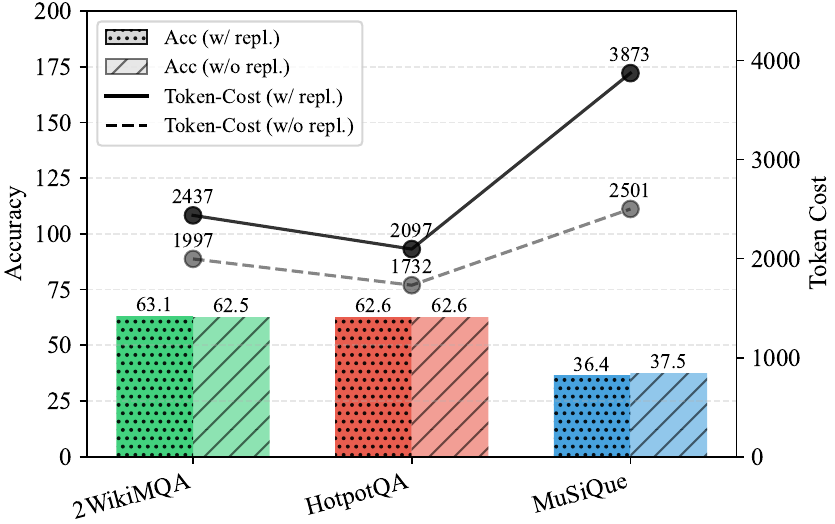}
        \caption{Comparison between two strategies: sampling w/ and w/o replacement.}
        \label{fig:strategy_compare}
    \end{minipage}
\end{figure*}

\subsection{Graph Reasoning Linearization}
\label{linear-g-reasoning}
Although query logic dependency graphs provide a structured approach to modeling complex queries, integrating RAG with logical reasoning remains challenging due to their inherent operational asymmetry: RAG assumes that queries are independent and self-contained, whereas reasoning requires the sequential processing of interdependent subproblems, with intermediate results composed and passed along logical dependencies.
A naïve integration of the two paradigms results in significant inefficiencies and semantic drift: each sub-query must be resolved in context, yet its formulation may depend on unresolved predecessors, creating a circular dependency in retrieval and generation.

Formally, let $ f_{\mathrm{RAG}}(p_i, \mathcal{C}_i) $ denote the retrieved answer for subproblem $ p_i $, given context $ \mathcal{C}_i $, and let $ \mathcal{R}(p_i) $ be a retrieval function producing $ \mathcal{C}_i $ from an external corpus $ \mathcal{K} $. Then the complete query resolution requires computing:
$$
A = \mathrm{Compose}\left( \{ f_{\mathrm{RAG}}(p_i, \mathcal{R}(p_i)) \}_{i=1}^n \right),
$$
where \textit{Compose} denotes the reasoning function that aggregates sub-results according to the logic encoded in the DAG $ G$.

However, due to the dependencies in $ G $, not all $ p_i $ can be processed in parallel. Specifically, the retrieval for $ p_i $ may depend on the answers to its parent nodes $ \mathrm{Pa}(v_i) $. This introduces a local directed constraint:
$$
\mathcal{C}_i = \mathcal{R}\left(p_i \,\middle|\, \{f_{\mathrm{RAG}}(p_j, \mathcal{C}_j) \}_{v_j \in \mathrm{Pa}(v_i)} \right),
$$
which prevents direct batching of retrievals. Without careful scheduling, this leads to recursive resolution that is inefficient and hard to optimize.

To address this, we exploit the acyclic nature of the dependency graph $ G $ and reduce the query resolution process to a two-step solution:

\textbf{Step 1: Topological Sort.} 
To efficiently schedule the resolution process, we perform a topological sort over the DAG $ G $, using a depth-first search traversal. This yields an ordered sequence of subproblems $ \langle p_{(1)}, p_{(2)}, \dots, p_{(n)} \rangle $, in which every subproblem appears after all its dependencies. This linearization respects the logical flow of reasoning and allows sequential resolution of subproblems in a single forward pass. The DFS-based topological sort has $O(V + E)$ time and space complexity.

\textbf{Step 2: Greedy Retrieval and Resolution.} We then iterate over the sorted subproblems, resolving each $ p_{(i)} $ greedily using the current context:
\begin{align}
        &\mathcal{C}_{(i)} = \mathcal{R}\left(p_{(i)} \,\middle|\, \{f_{\mathrm{RAG}}(p_{(j)}, \mathcal{C}_{(j)})\}_{j < i,\, v_{(j)} \in \mathrm{Pa}(v_{(i)})} \right),\\
&a_{(i)} = f_{\mathrm{RAG}}(p_{(i)}, \mathcal{C}_{(i)}).
\end{align}
This greedy approach eliminates recursion while maintaining logical consistency. Each context $ \mathcal{C}_{(i)} $ can be dynamically adapted based on previously retrieved and inferred sub-answers, enabling \textit{context-aware retrieval} and \textit{incremental reasoning} with linear complexity.

Overall, this DAG-guided reasoning framework transforms the complex interplay between retrieval and logic into a tractable pipeline. The explicit decomposition via topological sort and greedy resolution ensures efficiency, scalability, and faithfulness to the original query's reasoning structure.




\subsection{Graph and Context Pruning}
\label{sec:pruning}

While the topological sort linearizes the reasoning process, the overall inference cost remains non-trivial due to two factors: the growing size of accumulated context as subproblems are resolved, and the potential redundancy among subproblems with similar semantics or overlapping dependencies. To address this, we introduce a two-dimensional pruning mechanism that improves both computational efficiency and retrieval quality by reducing unnecessary context propagation and subproblem duplication.

\textbf{Context pruning via rolling memory.}  
As the RAG progresses through the topologically sorted subproblems, each retrieval step accumulates more contexts—text chunks retrieved from the knowledge base—which can overwhelm the LLM and introduce noise that impairs generation. To mitigate this, we maintain a compressed \textit{memory state}—a single text string that serves as a rolling summary of the most salient facts retrieved so far. After each subproblem $p_{(i)}$ is resolved, its retrieved context and answer $a_{(i)}$ are distilled via LLM-based summarization, and the resulting summary is incorporated into the memory. At each subsequent step, newly retrieved context is summarized oriented by the query and merged with the existing memory to form an updated memory:
\begin{equation}
\text{Mem}_{(i)} = \mathrm{Summarize}(\text{Mem}_{(i-1)} \cup \mathcal{R}(p_{(i)})),
\end{equation}
where $\mathcal{R}(p_{(i)})$ is the set of retrieved text chunks for subproblem $p_{(i)}$. This summarization-based pruning prevents context bloat while preserving only the most relevant information for downstream reasoning.

\textbf{Graph pruning via unified subquery generation.} 
Multiple subproblems in the DAG often share similar priorities or loosely coupled dependencies, such as leaf or sibling nodes in the topological order, representing parallelizable factoid sub-tasks. To reduce redundant retrievals and improve efficiency, we merge subproblems with the same topological rank into a single \textit{unified query}. For each step in the graph reasoning process (Section \ref{linear-g-reasoning}), let $S_{(i)} = \{p_{(j)} \mid \text{rank}(p_{(j)}) = \text{rank}(p_{(i)})\}_{j=i}^n$ denote the set of subproblems with the same rank as $p_{(i)}$. We construct a unified query:
\begin{equation}
q^{\mathrm{uni}}_{(i)} = \mathrm{Merge}(S_{(i)}),
\label{eq:sub_query_gen}
\end{equation}
and perform a single retrieval to obtain a unified context $\mathcal{C}^{\mathrm{uni}}_{(i)}$ and a response set $\{a_{(j)}\}_{p_{(j)} \in S_{(i)}}$. The process then advances to the next topological rank, skipping individual subproblem iterations. This graph pruning reduces retrieval operations, shortens the reasoning chain, and enhances consistency among semantically related subproblems.


By jointly applying context pruning and graph pruning, our framework achieves more efficient inference without compromising reasoning quality. The pruning process is tightly guided by the DAG and dynamically adapts to the evolving retrieval state, offering a scalable and context-aware solution for complex query answering.

\begin{table*}[t]\small
	\centering
	\begin{tabular}{lclccccccccc} 
	\toprule
 \multirow{2}{*}{\textbf{Type}} && \multirow{2}{*}{\textbf{Model}}&
    &\multicolumn{2}{c}{\textbf{HotpotQA}}
	& &\multicolumn{2}{c}{\textbf{2WikiMQA}} 
	&&\multicolumn{2}{c}{\textbf{MuSiQue}}\\
	 \cmidrule(lr){5-6} \cmidrule(lr){8-9} \cmidrule(lr){11-12}
	&& &&Str-Acc. &LLM-Acc. &&Str-Acc. &LLM-Acc. &&Str-Acc. &LLM-Acc.
        \cr \midrule 
    \multirow{4}{*}{\textbf{Direct Zero-shot LLM}} 
        && Llama3 (8b) &&  17.1 & 11.1 && 22.3 &	4.7 && 2.3 & 2.0 \\
        && Llama3 (13b) &&23.7& 20.1	&& 33.8 & 15.4 && 6.4 & 6.0\\
        && GPT-3.5-turbo && 31.5 & 35.4 && 24.0 & 22.0 && 7.9 & 10.9 \\
        && GPT-4o-mini &&38.7 & 36.3 &&26.4&24.3&&17.6&14.0
        \cr \midrule
    \multirow{3}{*}{\textbf{Vanilla RAG}} 
        && VanillaRAG (Top-$1$) &&38.4 & 48.6 && 34.8 & 37.3 &&13.2 & 18.5\\
        && VanillaRAG (Top-$3$) &&43.2 & 53.1 && 43.0 & 42.0 &&20.3 & 23.6\\
        && VanillaRAG (Top-$5$) &&44.1 & 53.9 &&46.7 & 45.6 && 21.0 & 23.6
        \cr \midrule 
    \multirow{6}{*}{\textbf{Graph-based RAG}} 
        && KGP&& 46.4& 57.1 && 47.5& 43.7&& 23.3& 27.5 \\
        && G-retriever  && 28.5&40.9 && 26.7& 35.7 && 9.1 & 15.6\\ 
        && RAPTOR && 48.1 & 57.8 && 47.7&45.9 && 25.2 &29.1 \\
        && GraphRAG  &&39.6& 45.2 && 46.3&43.3 &&16.5&23.1\\
        && LightRAG  &&47.8 &57.7 &&43.1 &36.3 &&18.1 &19.4\\
        && HippoRAG  &&53.7 &55.6 &&47.7 & 47.2 &&24.9 &30.1 \\
        && HippoRAG2&&\textbf{56.7} &61.9&& 50.0&47.1&&27.0&32.6
        \cr \midrule
    \multirow{1}{*}{\textbf{Ours}} 
        && LogicRAG &&54.8&\textbf{62.6}&&\textbf{64.7}&\textbf{62.5}&&\textbf{30.4}&\textbf{37.5}
        \cr 
        \bottomrule
	\end{tabular}
    	\caption{Question answering accuracy across benchmark datasets.} 
    \label{tab:main_results}
\end{table*}

\subsection{Sampling Strategy}

We have developed a logic-guided RAG model that operates in an iterative, agentic manner. However, such systems often encounter a ``hesitation'' issue, wherein the language model, faced with uncertainty, repeatedly generates similar subqueries. This behavior impedes progress in resolving downstream subproblems and leads to inefficiencies in both computation and information gathering.

During the subquery generation step in Equation~\ref{eq:sub_query_gen}, we consider two strategies for iterating over subproblems: (1) \textit{sampling with replacement}, where the current batch of subproblems \( S_{(i)} \) is retained after generating the unified subquery \( q^{\text{uni}}_{(i)} \), allowing the language model to decide whether to proceed to the next batch \( S_{(i+1)} \); and (2) \textit{sampling without replacement}, where \( S_{(i)} \) is removed after generating \( q^{\text{uni}}_{(i)} \), enforcing forward progression to \( S_{(i+1)} \). The first strategy, commonly adopted in iterative frameworks~\citep{fu2023specializing,yang2024rag,li2021graph}, is prone to stalling at intermediate stages, producing near-duplicate subqueries until the maximum number of retrieval iterations is reached.

Figure~\ref{fig:sq_sim} presents the similarity matrix of subqueries generated under the first strategy. The heatmap indicates an increasing tendency toward repetition as iterations progress. Figure~\ref{fig:strategy_compare} compares token consumption and answer accuracy under both strategies. Sampling without replacement consistently yields lower per-question token cost across three datasets while maintaining comparable answer quality. We attribute this to two factors: (1) forcing progression prevents the model from lingering on uncertain subproblems and encourages the assimilation of broader knowledge, and (2) avoiding redundant subquery generation leads to more efficient inference.

Motivated by these findings, we adopt sampling without replacement as the default strategy to ensure efficient and effective multi-step retrieval and reasoning.

\section{Experiments}

\subsection{Experiment Setup}

\textbf{Datasets.} We evaluate our method's RAG capabilities using three multi-hop question-answering benchmarks: MuSiQue (answerable)~\cite{trivedi2022MuSiQue}, 2WikiMultiHopQA (abbreviated as 2WikiMQA)~\cite{2wikimqa}, and HotpotQA~\cite{yang2018hotpotqa}. To manage experimental costs, we follow HippoRAG~\cite{HippoRAG} by extracting 1,000 questions from each dataset's validation set. Additionally, we adopt the approach of IRCOT~\cite{IRCOT} and HippoRAG~\cite{HippoRAG} to gather all candidate passages, including both supporting and distractor passages, from the selected questions to create a retrieval corpus for each dataset. Dataset details are presented in Table~\ref{dataset_stat}.

\begin{table}[h]\small
    \centering
    \begin{tabular}{lccc}
        \toprule
        & MuSiQue & 2WikiMQA & HotpotQA \\
        \midrule
        \# of Passage & 11,656 & 6,119 & 9,221 \\
        \# of Query type &3&4&2\\
        \bottomrule
    \end{tabular}
    \caption{Dataset statistics.}
    \label{dataset_stat}
    \vspace{-1em}
\end{table}
\textbf{Baselines.} We compare against several baselines: (i) Zero-shot LLM Inference (llama3(8b), llama3(13b)~\cite{llamaindex_website}, gpt3.5-turbo, gpt-4o-mini~\cite{achiam2023gpt}), (ii) vanilla retrieval augmented generation (top1, top3, top5), (iii) graph-based retrieval augmented generation (KGP~\cite{wang2024knowledge}, RAPTOR~\cite{sarthi2024raptor}, G-retriever~\cite{he2024g}, GraphRAG~\cite{edge2024local}, LightRAG~\cite{guo2024lightrag}, HippoRAG~\cite{HippoRAG}, HippoRAG2~\cite{hipporag2}). For those who provide both single-step and multi-step options, we use their multi-step version for its general superior performance. 


\textbf{Metrics.} We evaluate end-to-end QA performance on the datasets using two metrics: \textit{string-based accuracy}, which computes whether the gold answer is included in the generated answer after normalizing them to lowercase words; and \textit{LLM-based accuracy}, which lets an LLM decide whether the generated answer correctly matches the gold answer.

\begin{figure*}[t]
    \centering
\includegraphics[width=.95\linewidth]{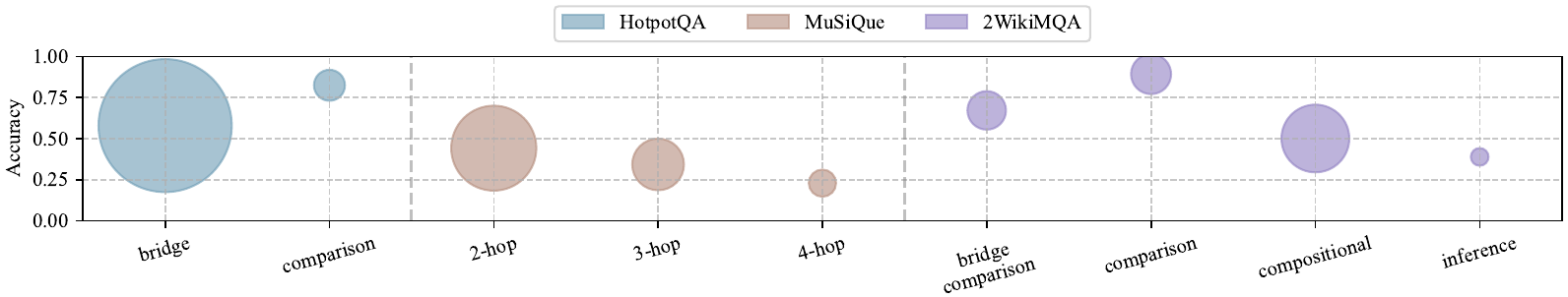}
    \caption{Distribution of accuracy across question types. Each ball represents a question type, with the y-axis position indicating its accuracy and the radius reflecting its proportion in the dataset.}
    \label{fig:q_type_dist}
\end{figure*}

\begin{figure*}[h]
    \centering
    \includegraphics[width=\linewidth]{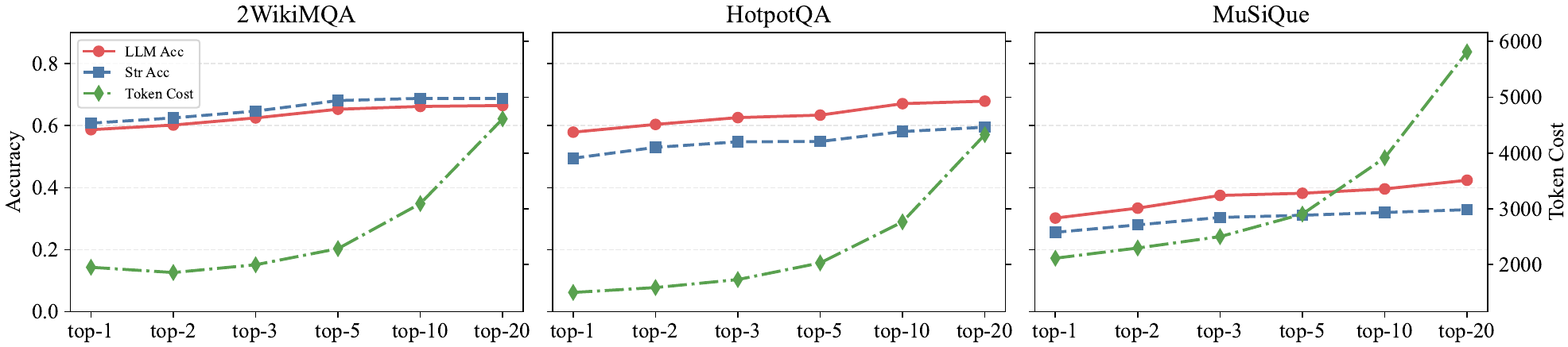}
    \caption{Pareto frontier of efficiency versus effectiveness for different k values.}
    \label{fig:topk-effect}
\end{figure*}

\begin{table}[!h]\small
    \centering

    \begin{tabular}{@{}ll@{}}
        \toprule
        Component            & Specification                          \\
        \midrule
        GPU       & NVIDIA GeForce RTX 3090                \\
        CPU                    & Intel Xeon (R) Silver (R) 4214R CPU 2.40 @ GHz \\
        \bottomrule
    \end{tabular}
        \caption{Machine configuration.}
    \label{config}
\end{table}

\textbf{Implementation details.} To ensure fair comparison, we use the same embedding models (sentence-transformers/all-MiniLM-L6-v2) for all algorithms. The $k$ is set to 3 for each top-$k$ retrieval. All the RAG methods use the same large language models used for generation and evaluation, which is gpt-4o-mini. Hardware configurations of the experiments are presented in Table~\ref{config}.

\subsection{Main Results}

Table~\ref{tab:main_results} reports the end-to-end question answering accuracy across three benchmarks, comparing LogicRAG against baselines. Metrics include \textit{string-based accuracy} and \textit{LLM-based accuracy}. The results lead to several key observations:

\textbf{Zero-shot LLM inference.} Direct prompting of LLMs without any retrieval yields poor performance across all datasets. Even strong models like GPT-4o-mini achieve only 38.7\% string accuracy on HotpotQA and significantly lower on 2WikiMQA and MuSiQue, underscoring the necessity of retrieval augmentation for knowledge-intensive questions.

\textbf{Vanilla RAG.} Integrating corpus retrieval substantially enhances performance. Increasing $k$ consistently improves accuracy across all datasets. However, the performance gains from larger $k$ diminish at higher values, revealing that vanilla RAG's limitation lies in its bottleneck: reasoning over multi-hop dependencies.

\textbf{Graph-based RAG.} Methods that explicitly model relationships among text chunks via graph structures further enhance performance. HippoRAG2 achieves the best results among existing methods, reaching 56.7\% string accuracy on HotpotQA and 27.0\% on MuSiQue. However, these methods remain limited in integrating logic constraints or leveraging structural reasoning during retrieval.

\textbf{LogicRAG} outperforms all baselines by a significant margin across all datasets. Notably, LogicRAG achieves 64.7\% string accuracy on 2WikiMQA, a +14.7\% absolute improvement over the next best baseline (HippoRAG2), and boosts GPT-based accuracy to 62.5\%. Similarly, on MuSiQue, LogicRAG reaches 30.4\% string accuracy and 37.5\% LLM accuracy, outperforming HippoRAG2 by +3.4\% and +4.9\%, respectively. These results highlight the effectiveness of structured, logic-guided reasoning in enhancing retrieval and QA quality in multi-hop settings.

\subsection{Performance Breakdown by Question Type}

Figure~\ref{fig:q_type_dist} presents the distribution of LogicRAG’s accuracy across question types with a ball plot. This visualization allows us to assess the proportion of query types and their impact on model performance.

On HotpotQA, LogicRAG demonstrates strong performance on \texttt{comparison} questions, achieving an accuracy of 83\%, significantly higher than the 58\% accuracy on the more prevalent \texttt{bridge} questions. This suggests that LogicRAG is particularly effective in tasks that require identifying and comparing specific facts, while performance on bridge-type questions—often requiring entity chaining—is more susceptible to retrieval errors or reasoning drift.

In MuSiQue and 2WikiMQA, we observe more diverse reasoning patterns. For MuSiQue, accuracy consistently declines as the number of reasoning hops increases. This trend underscores the growing complexity of multi-hop reasoning and the challenge of maintaining coherent retrieval chains. In 2WikiMQA, LogicRAG excels at \texttt{comparison} (89\%) and \texttt{bridge-comparison} (67\%) questions. In contrast, it shows lower performance on \texttt{inference} questions (39\%) and moderate performance on \texttt{compositional} questions (50\%), which represent the largest category (44.4\%). Given both the prevalence and the difficulty of the compositional type, improving model capability in handling compositional reasoning would likely yield substantial gains in overall performance on 2WikiMQA and represents a valuable direction for future research.

\subsection{Efficiency-Effectiveness Trade-off of Top-$k$}

Figure~\ref{fig:topk-effect} illustrates the trade-off between the overall performance and efficiency of LogicRAG of varied top-$k$ selection. We plot both accuracy (left y-axis) alongside the corresponding token cost (right y-axis) of the overall RAG task. This figure serves as a Pareto frontier analysis, highlighting how $k$ impacts answer quality and computational overhead.

Across all datasets, both accuracy metrics generally improve as $k$ increases, but with diminishing returns beyond top-$3$ or top-$5$. For example, on 2WikiMQA, increasing $k$ from $1$ to $5$ leads to a notable gain in string accuracy, while gains from top-$10$ to top-$20$ are marginal. A similar trend holds for HotpotQA, where LLM accuracy saturates near top-$10$. On MuSiQue, improvements are more gradual, reflecting the inherent difficulty and more dispersed knowledge distribution in this dataset.

However, increasing $k$ incurs a steep rise in token cost, particularly in MuSiQue, where the average cost exceeds 6000 tokens at top-$20$. This growing cost makes high $k$ values less practical in real-world applications where latency or budget constraints are critical. Taken together, these results suggest that moderate values (e.g., $k=3$ or $k=5$) offer the best balance between effectiveness and efficiency. This insight also motivates our context pruning with rolling memory that summarizes salient facts during iterative RAG to avoid incremental context accumulation.

\begin{figure}[h]
    \centering
    \includegraphics[width=0.8\linewidth]{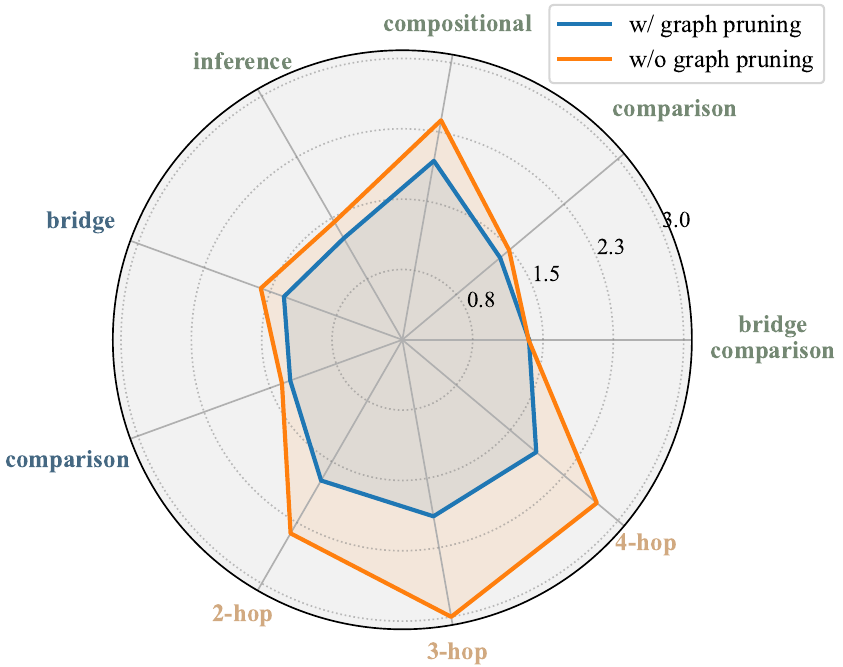}
    \caption{Impact of graph pruning, by comparing the average number of retrieval rounds with and without graph pruning. }
    \label{fig:g_pruning}
    \vspace{-1em}
\end{figure}

\subsection{Impact of Graph Pruning}

We evaluate the impact of graph pruning by comparing the average number of retrieval rounds across different query types, as shown in Figure \ref{fig:g_pruning}. Overall, graph pruning consistently reduces the number of retrieval steps across all categories, confirming its effectiveness in eliminating redundant subproblem resolution and promoting efficient inference. Notably, compositional and multi-hop questions—typically requiring deeper reasoning—benefit most from pruning due to their higher potential for semantic overlap across subqueries.

Interestingly, while multi-hop questions (2-hop to 4-hop) generally demand more retrieval rounds, the increase is not strictly monotonic. In particular, 4-hop questions exhibit fewer retrieval rounds on average than 3-hop questions. Our investigation reveals that this arises from a behavioral tendency of the LLM: \textit{when faced with long, complex queries, it sometimes becomes prematurely confident and produces an answer based on partial information.} This mirrors a well-known human cognitive bias—those with less insight may display higher confidence—highlighting the challenge of trust calibration in LLM-driven reasoning.

\begin{table}[h]\small
\centering
\begin{tabular}{l c c}
\toprule
\textbf{Method} & \textbf{Avg. Time (s)} & \textbf{Avg. Token} \\
\midrule
ZeroShot     & 5.88  & 216.2   \\
VanillaRAG   & 4.28  & 489.7  \\
\midrule
G-retriever  & 12.50 & 1000.0  \\
KGP          & 70.72 & 11097.8 \\
Raptor       & 5.79  & 2568.0  \\
GraphRAG     
& 13.05 & 4699.8  \\
LightRAG     & 35.14 & 5730.6  \\
HippoRAG     & 6.30  & 2608.8  \\
HippoRAG2    & 5.89  & 2809.2  \\
\midrule
LogicRAG     & 9.83 & 1777.9  \\
\bottomrule
\end{tabular}
\caption{Query answering efficiency on 2WikiMQA.}
\label{tab:qa-efficiency}
\vspace{-1em}
\end{table}

\subsection{Query-time Efficiency Comparison}

Table~\ref{tab:qa-efficiency} reports the query-time efficiency of various multi-hop QA models on 2WikiMQA, measured by average latency and token usage during the retrieval and generation stages. Notably, this comparison excludes the cost of graph construction, which is required by all graph-based baselines. These methods typically rely on costly preprocessing to build knowledge graphs, including steps like named entity recognition, open information extraction, and offline graph generation. The constructed graphs are then used to summarize and index the corpus. This preprocessing often takes tens to hundreds of minutes and can incur substantial additional token usage, as shown in Table~\ref{fig:graph-construction-cost}.

By contrast, LogicRAG eliminates the need for any offline graph construction, yet still maintains competitive query-time performance with a modest latency and moderate token usage. While flat retrieval baselines like VanillaRAG and ZeroShot achieve slightly faster response times, they lack the structured reasoning capabilities enabled by LogicRAG. Taken together, these results suggest that LogicRAG offers a cost-effective and practical solution for multi-hop QA, particularly in deployment scenarios where preprocessing and knowledge update overhead is a bottleneck.


\section{Conclusion}

In this work, we present LogicRAG, a novel retrieval-augmented generation framework that dynamically leverages reasoning structures for complex query resolution. By decomposing queries into subproblems and modeling their dependencies as a directed acyclic graph, LogicRAG enables logic-aware and efficient multi-step retrieval. 
This principled framework offers a practical and scalable solution for knowledge-intensive complex question answering with large language models.

\section*{Acknowledgement}
The work described in this paper was fully supported by a grant from the Innovation and Technology Commission of the Hong Kong Special Administrative Region, China (Project No. ITS/263/24FP).

\bibliography{logicrag}

\cleardoublepage
\appendix

\onecolumn

\section{Supplementary Material}
\label{app:myappendix}  

\newtcolorbox{questionbox}{
    colback=blue!3!white,
    colframe=blue!70!black!30!gray,
    fonttitle=\bfseries\large,
    title=Question,
    sharp corners,
    boxrule=1pt,
    left=6pt, 
    top=4pt,
    bottom=4pt,
    enhanced,
    before upper={\parindent0pt\noindent}
}

\newtcolorbox{answerbox}{
    colback=green!3!white,
    colframe=green!50!black!30!gray,
    fonttitle=\bfseries\large,
    title=Answer,
    sharp corners,
    boxrule=1pt,
    left=6pt,
    top=4pt,
    bottom=4pt,
    enhanced,
    before upper={\parindent0pt\noindent}
}

\subsection{A. Pseudo Code}

\begin{algorithm}[hpt]
\caption{LogicRAG: Logic-Aware Retrieval-Augmented Generation}
\label{alg:logicrag}
\begin{algorithmic}[1]
\State \textbf{Input:} Query $Q$, LLM, embedding model $\theta$, knowledge corpus $K$
\State \textbf{Output:} Final answer $A$
\State Use LLM to decompose $Q$ into subproblems $P = \{p_1, \dots, p_n\}$
\State Initialize DAG $G = (V, E)$ with $V = \{v_i \leftrightarrow p_i\}$ and $E = \emptyset$
\State Use LLM to infer logical dependencies among $P$ and populate $E$
\State Topologically sort $G$ to obtain ranks $\text{rank}(p_i)$ for all $p_i$
\State Initialize rolling memory $\text{Mem}^{(0)} \gets \emptyset$
\For{each rank $r$ in ascending topological order}
    \State Let $S^{(r)} = \{p_i \mid \text{rank}(p_i) = r\}$ \Comment{Subproblems at same level}
    \State Construct unified query $q^{(r)} = \texttt{Merge}(S^{(r)})$ using LLM
    \State Retrieve documents $C^{(r)} = R(q^{(r)}; \theta, K)$ \Comment{Unified retrieval}
    \State Summarize $C^{(r)}$ with respect to $Q$ and $\text{Mem}^{(r-1)}$ to get $\text{Mem}^{(r)}$
    \For{each $p_i \in S^{(r)}$}
        \State Use LLM to resolve $p_i$ using $\text{Mem}^{(r)}$ and store intermediate answer $a_i$
    \EndFor
  \If{LLM identifies a new unresolved subproblem $p_{n+1}$}
        \State Add new node $v_{n+1}$ to $V$ and edge(s) to $E$ if needed
        \State Append $p_{n+1}$ as a new rank after the current rank sequence
    \EndIf
\EndFor
\State Generate final answer $A = \texttt{Compose}(\{a_i\})$ using LLM
\State \textbf{Return} $A$
\end{algorithmic}
\end{algorithm}

\subsection{B. Related Work}

\textbf{Graph-based RAG} methods explicitly structure textual corpora into graphs to support multi-hop reasoning. For example, hierarchical-graph methods like GraphRAG~\cite{edge2024local} and RAPTOR~\cite{sarthi2024raptor} organize knowledge into multi-level document graphs (via community detection or recursive tree summarization) for coarse-to-fine retrieval. LightRAG~\cite{guo2024lightrag} similarly builds a dual-level, graph-augmented index to enable efficient, incrementally updatable retrieval. Inspired by hippocampal memory models, HippoRAG~\cite{HippoRAG} and HippoRAG2~\cite{hipporag2} constructs a knowledge graph to index text chunks, and employ Personalized PageRank to perform one-shot multi-hop search. These graph-based approaches can employ off-the-shelf graph reasoning algorithms~\cite{difflogic, neusymea, zhou2025taming, liu2023rsc, bei2023non, chen2024graphcrosscorelated,li2023gslb} for enhancing retrieval accuracy. However, they rely on pre-built knowledge graphs or indexes. In contrast, LogicRAG eschews any fixed graph: it dynamically constructs a logic dependency graph of intermediate subqueries at inference time and uses a topological ordering to guide retrieval in a logically coherent sequence. LogicRAG also prunes redundant graph branches and context passages at each step to reduce overlap and token usage, enabling efficient multi-hop reasoning without a precomputed graph structure.

\textbf{Reasoning-enhanced RAG} methods instead interleave retrieval with the model’s own reasoning process. Think-on-Graph~\cite{ma2024think-on-graph-2.0} alternates between using a static knowledge graph and text retrieval, leveraging entity links to deepen contextual retrieval iteratively. Likewise, RRP~\cite{xiao2025reliablereasoningpathdistilling} distills guidance for LLM reasoning on well-established knowledge graphs. IRCoT~\cite{IRCOT} interleaves retrieval with a chain-of-thought: it uses the LLM’s intermediate reasoning steps to decide what to retrieve next, feeding retrieved facts back into the CoT. Likewise, IM-RAG~\cite{im-rag} trains the LLM to generate iterative ``inner monologue'' queries over multiple rounds of retrieval, refining its queries via reinforcement learning. Recent LAG~\cite{xiao2025laglogicaugmentedgenerationcartesian} enhances problem solving by problem decomposition and logic augmentation from a Cartesian perspective. These methods improve multi-step QA by tightly coupling retrieval with reasoning steps, but they still depend on fixed resources (like a static KG) or predetermined retrieval schedules. LogicRAG, by contrast, constructs and updates its reasoning DAG adaptively during inference. It schedules subgoals by topologically sorting the DAG, effectively planning retrieval in dependency order, and prunes irrelevant branches and context on the fly. This dynamic, adaptive planning lets LogicRAG perform flexible multi-step reasoning efficiently, without any static graph or fixed retrieval plan.

\subsection*{C. Notations}

Please refer to Table~\ref{tab:notation}.

\begin{table}[hpt]
\centering
\begin{tabular}{ll}
\toprule
\textbf{Symbol} & \textbf{Description} \\
\midrule
$Q$ & Input query \\
$A$ & Final answer generated by the RAG system \\
$K$ & External knowledge corpus \\
$C$ & Retrieved context for the query $Q$ \\
$R(\cdot)$ & Retrieval function mapping queries to relevant contexts from $K$ \\
$P = \{p_1, \dots, p_n\}$ & Set of subproblems decomposed from the input query $Q$ \\
$G = (V, E)$ & Query logic dependency graph (a DAG) modeling subproblem dependencies \\
$V = \{v_1, \dots, v_n\}$ & Node set of $G$, where $v_i$ corresponds to subproblem $p_i$ \\
$E \subseteq V \times V$ & Directed edge set representing logical dependencies among subproblems \\
$\mathrm{Pa}(v_i)$ & Set of parent nodes of $v_i$ in the DAG $G$ \\
$f_{\mathrm{RAG}}(p_i, C_i)$ & RAG model output for subproblem $p_i$ with context $C_i$ \\
$a_i$ & Answer generated for subproblem $p_i$ \\
$\mathrm{Mem}^{(i)}$ & Rolling memory at step $i$, containing  summarized prior context \\
$\mathrm{rank}(p_i)$ & Topological rank of subproblem $p_i$ in DAG $G$ \\
$S^{(i)}$ & Set of subproblems with the same  rank $\mathrm{rank}(p_i)$ \\
$q^{\mathrm{uni}}_i$ & Unified query constructed from $S^{(i)}$ \\
$C^{\mathrm{uni}}_i$ & Retrieved context for the unified query $q^{\mathrm{uni}}_i$ \\
\bottomrule
\end{tabular}
\caption{Notation Table}
\label{tab:notation}
\end{table}

\subsection*{D. Case Study}

We show a case of a three-hop question from the MuSiQue dataset, to demonstrate how to identify supporting knowledge for the question by multiround retrieval with reasoning. 

This example demonstrates a complex multi-hop question requiring geopolitical knowledge, temporal reasoning, and indirect entity resolution. The question refers obliquely to the Soviet Union through the phrase \textit{“the country where, despite being headquartered in the nation called the nobilities commonwealth, the top-ranking Warsaw Pact operatives originated,”} which demands recognizing that the \textit{“nobilities commonwealth”} is a historical reference to the Polish–Lithuanian Commonwealth, and that while the Warsaw Pact was signed in Warsaw, Poland, it was dominated by the USSR. Thus, the country in question is the Soviet Union, making the Tripartite discussions involve Britain, France, and the USSR.

To answer the temporal aspect of the question, the model must locate when discussions involving these three parties began. \textit{RetrievalBox3} provides a precise temporal signal: \textit{“In mid-June, the main Tripartite negotiations started.”} This date, paired with the correct identification of the USSR, allows for an accurate answer—June. The example illustrates the model’s ability to resolve indirect references, integrate multi-document evidence, and infer temporal details grounded in historical context.

\newtcolorbox{retrievalbox}{
    colback=orange!3!white,
    colframe=orange!80!black!80!,
    fonttitle=\bfseries\large,
    title=Retrieval Round 1,
    sharp corners,
    boxrule=1pt,
    left=6pt,
    top=4pt,
    bottom=4pt,
    enhanced,
    before upper={\parindent0pt\noindent}
}

\newtcolorbox{retrievalbox2}{
    colback=orange!3!white,
    colframe=orange!80!black!80!,
    fonttitle=\bfseries\large,
    title=Retrieval Round 2,
    sharp corners,
    boxrule=1pt,
    left=6pt,
    top=4pt,
    bottom=4pt,
    enhanced,
    before upper={\parindent0pt\noindent}
}

\newtcolorbox{retrievalbox3}{
    colback=orange!3!white,
    colframe=orange!80!black!80!,
    fonttitle=\bfseries\large,
    title=Retrieval Round 3,
    sharp corners,
    boxrule=1pt,
    left=6pt,
    top=4pt,
    bottom=4pt,
    enhanced,
    before upper={\parindent0pt\noindent}
}

\begin{questionbox}
What month did the Tripartite discussions begin between Britain, France, \textbf{and the country where, despite being headquartered in the nation called the nobilities commonwealth, the top-ranking Warsaw Pact operatives} originated?
\end{questionbox}

\begin{retrievalbox}
The Warsaw Pact, formally known as the Treaty of Friendship, Cooperation and Mutual Assistance, was a collective defence treaty signed in Warsaw, Poland among the Soviet Union and seven Soviet satellite states of Central and Eastern Europe during the Cold War. The Warsaw Pact was the military complement to the Council for Mutual Economic Assistance (CoMEcon), the regional economic organization for the socialist states of Central and Eastern Europe. The Warsaw Pact was created in reaction to the integration of West Germany into NATO in 1955 per the London and Paris Conferences of 1954, but it is also considered to have been motivated by Soviet desires to maintain control over military forces in Central and Eastern Europe.
\end{retrievalbox}

\begin{retrievalbox2}
The eight member countries of the Warsaw Pact pledged the mutual defense of any member who would be attacked. Relations among the treaty signatories were based upon mutual non-intervention in the internal affairs of the member countries, respect for national sovereignty, and political independence. However, almost all governments of those member states were indirectly controlled by the Soviet Union.
\end{retrievalbox2}

\begin{retrievalbox3}
By the end of May, drafts were formally presented. In mid-June, the main Tripartite negotiations started. The discussion was focused on potential guarantees to central and east European countries should a German aggression arise. The USSR proposed to consider that a political turn towards Germany by the Baltic states would constitute an "indirect aggression" towards the Soviet Union. Britain opposed such proposals, because they feared the Soviets\' proposed language could justify a Soviet intervention in Finland and the Baltic states, or push those countries to seek closer relations with Germany. The discussion about a definition of "indirect aggression" became one of the sticking points between the parties, and by mid-July, the tripartite political negotiations effectively stalled, while the parties agreed to start negotiations on a military agreement, which the Soviets insisted must be entered into simultaneously with any political agreement.
\end{retrievalbox3}

\begin{answerbox}
\textbf{June}
\medskip
\textit{Supporting Evidence:}
\begin{itemize}[leftmargin=*,nosep]
    \item Historical records state: "\textbf{In mid-June}, the main Tripartite negotiations started"
    \item Contextual factors:
    \begin{itemize}[leftmargin=*]
        \item The "nobilities commonwealth" refers to the USSR (headquartered in Moscow)
        \item Warsaw Pact operatives were predominantly Soviet-controlled
        \item Discussions focused on Central/Eastern European security guarantees
    \end{itemize}
\end{itemize}
\end{answerbox}

\end{document}